# Performance Evaluation of Bitstring Representations in a Linear Genetic Programming Framework

Clyde Meli[1[0000-0003-3551-862X]], Vitezslav Nezval[1], Zuzana Kominkova Oplatkova[2[0000-0001-8050-162X]], Victor Buttigieg[3[0000-0002-8875-1348]], Anthony Spiteri Staines[1[0000-0002-2747-9766]]

Dept. Computer Information Systems
University of Malta
{clyde.meli@um.edu.mt,vnez@cis.um.edu.mt,tony.spiteri-staines@um.edu.mt}
[2]Faculty of Applied Informatics, Dept. of Informatics and Artificial Intelligence
Tomas Bata University in Zlín
Czech Republic
oplatkova@utb.cz
[3]Dept. of Communications & Computer Engineering
University of Malta
victor.buttigieg@um.edu.mt

**Abstract:** Different bitstring representations offer different performance computations. This work describes three different bitstring representations: i) std::bitset, ii) Boost::dynamic_bitset, and iii) a custom direct implementation, written in the C++ programming language. Their performance is benchmarked in the context of concatenation in a Linear Genetic Programming (LGP) system. The benchmarks were conducted on three platforms—i) macOS, ii) Linux, and iii) Windows (MSYS2)—to assess platform-specific performance variations. The results show that the custom direct implementation delivers the fastest performance on ii) Linux and iii) Windows, while std::bitset performs best on i) MacOS. Boost::dynamic_bitset, although consistently slower, is still a viable and flexible choice. The findings highlight the influence of compiler optimisations and system architecture on performance, providing actionable insights for selecting the optimal method based on platform and application requirements. Keywords: C++, bitstring, linear genetic programming (LGP), benchmarking, C++ Standard Template Library (STL), Boost C++ Libraries

## 1    Problem Definition

This paper examines three different implementations for a bitstring data structure, in the C++ programming language written for a Linear Genetic Programming (LGP) System[1].

We compare the Boost library's[2] implementation of bitstring, a direct implementation using an array of ints (integer array), as well as the use of std::bitset from the standard C++ library. A bitset is an array of bool (Boolean array). The main reason is that sometimes anecdotally[1] it is suggested that an array of ints is faster than the use of std::bitset in the library itself. However this might not be the case in all scenarios and conditions.

Merelo et al[3] looked at the benchmarking of languages for evolutionary algorithms. Here the emphasis is on the improvements which can arise from the specific ways in which algorithms can be implemented. The authors concentrated on bit mutation and two-point crossover operations in thirteen programming languages. PHP proved to be unusually faster than C++ in their implementation which may suggest a problem with the C++ implementation.

O'Dwyer[4] wrote about bit-vector manipulation in C++. For small N<64 use of unsigned long long was recommended. For N<128 use of _uint128 was recommended. Otherwise, std::bitset<N> which is statically sized and stack allocated. A benchmark is provided, but we are told to take it with a "grain of salt" as these were produced in a Virtual Machine (VM) at godbolt.org.

Knauth et al[5] investigated the bit-reverse permutation and benchmarked five different optimised implementations with the use of CUDA[6].

Gustaffson[7] investigated the efficiency of bitstring binary pattern matching and construction in Erlang. No multiple runs were performed.

Pieterse et al[8] benchmarked C++ Bit-vector implementations. They compared std::vector<bool> (or std::vector<char> which at times they mention), boost::dynamic_bitset, Qt::QBitArray from QT software and Bitmagic's bm::bvector<>. They did not compare a native direct C++ implementation or the standard library std::bitset. 10 runs were utilised for each test and -O3 optimisations were used. boost::dynamic_bitset came out on top for execution speed. For memory efficiency std::vector<char> was found to be the best.

## 2    Methods

### 2.1    Hardware Equipment Used

The experiments were performed on two machines, with three operating systems (OS) as follows:

-    Desktop PC with dual boot Windows 10 and Void Linux
-    Laptop MacBook with Sonoma.

Full tech specs are given in Appendix 1.

---

[1] See https://stackoverflow.com/questions/30295174/what-is-the-performance-of-stdbitset

## 2.2 Methodology for Benchmarking the Bitstring Class

The three types of strings (direct implementation string, boost::dynamic_bitset string and std::bitset string) used for benchmarking through Google Benchmark[9] were implemented using the same Bitstring programming class in C++, and compiled with the GCC 15 C++ Compiler. The specific string type was selected at compile time through a C++ preprocessor directive. No virtualisation was performed to avoid skewed results. By direct it is meant that an array of Booleans (Bool data type) is used to implement the chromosome structure. Std::bitset means the STL std::bitset container is used to implement it. Boost::dynamic_bitset means the Boost library dynamic bitset container is used. All three algorithms have O(N) complexity for an N-bit string, differing only in constant factors i.e. packing efficiency and small fixed overheads. std::bitset and Boost::dynamic_bitset would use less memory footprint by packing 1 bit per element. The C++20 language standard was used. It is used in the Linear Genetic Programming (LGP) system developed for [1], [10].

The goal of the test was to determine the relative efficiency of operations on the three type of strings running on three different OSes. To evaluate this through benchmarking we ran a test 15 times, timing the Bitstring class. In the test two bitstrings are generated and concatenated using the plus operator.

The process iteratively generates all possible combinations of these two bitstrings. Initially, both bitstrings are assigned a value of zero. For each iteration, the two 4-bit values are concatenated to form an 8-bit composite bitstring, subsequently this is converted into its corresponding decimal representation.

Two nested loops are utilised. The outer loop ranges from 0-15, representing the first 4-bit value. The inner loop ranges from 0 to Jend, where Jend varies from 1 to 15. Each concatenated 8-bit value produced within these iterations serves as an input instance for the benchmarking procedure, enabling performance analysis of bitstring concatenation operations.

Both elapsed and CPU time are reported by Google Benchmark. Elapsed time is the total real time taken which might include time while the process was swapped out while something else was running. As a result, elapsed time should not be used for benchmarking and CPU time is what will be compared. Elapsed time and CPU time can be estimated as follows in Google Benchmark.

**Elapsed Time Per Iteration = (End Time - Start Time) / (Number of Iterations)**

**CPU Time per Iteration = (CPU End Time – CPU Start Time) / (Number of Iterations)**

It reports the mean, median, standard deviation and coefficient of variation of the runs. It reports on the iterations performed, which are the number of times a benchmark has been run until a statistically stable value has been reached by Google Benchmark. This feature eliminates statistical noise and generates a fair comparison. Note that this happens in a single run in our program.

The command **cpupower frequency-set --governor performance** was used to disable CPU scaling under Linux as root.

## 3    Results

Several experiments were carried out using multiple system configurations. The bench-marking results are shown in Tables 1-6, which were obtained using these standard conditions, three operating systems – MacOS, Windows 10 (MSYS2) and Void Linux. Tables 1 to 3 show the results by implementation, whereas tables 4 to 6 show them by OS. CPU time is given in *ms*. Iterations are always 500.

Table 1. Direct Implementation type of string

| Macos | | | Linux | | | Win10 | |
|---|---|---|---|---|---|---|---|
| | Cpu time | | | Cpu time | | | Cpu time |
| Mean | 2.09294 | | Mean | 1.28494 | | Mean | 1.3221 |
| Median | 2.09056 | | Median | 1.28477 | | Median | 1.32597 |
| Stddev | 0.013083 | | Stddev | 0.0011873 | | Stddev | 0.011807 |
| cv | 0.012502 | | Cv | 0.0018481 | | Cv | 0.008931 |

Table 2. Boost Implementation type of string

| Macos | | | Linux | | | Win10 | |
|---|---|---|---|---|---|---|---|
| | Cpu time | | | Cpu time | | | Cpu time |
| Mean | 1.44433 | | Mean | 1.40608 | | Mean | 1.44832 |
| Median | 1.39833 | | Median | 1.40581 | | Median | 1.44231 |
| Stddev | 0.197199 | | Stddev | 0.0020054 | | Stddev | 0.014543 |
| cv | 0.273065 | | Cv | 0.0028525 | | Cv | 0.010041 |

Table 3. std::bitset Implementation type of string

| Macos | | | Linux | | | Win10 | |
|---|---|---|---|---|---|---|---|
| | Cpu time | | | Cpu time | | | Cpu time |
| Mean | 1.29317 | | Mean | 1.32964 | | Mean | 1.36632 |
| Median | 1.28691 | | Median | 1.32958 | | Median | 1.3587 |
| Stddev | 0.018745 | | Stddev | 0.0007834 | | Stddev | 0.01554 |
| cv | 0.028991 | | Cv | 0.0011783 | | Cv | 0.011374 |

Table 4. Macos comparison

| Direct | | | Boost | | | std::bitset | |
|--------|----------|---|-------|----------|---|-------------|----------|
| | Cpu time | | | Cpu time | | | Cpu time |
| Mean | 2.09294 | | Mean | 1.44433 | | Mean | 1.29317 |
| Median | 2.09056 | | Median | 1.39833 | | Median | 1.28691 |
| Stddev | 0.013083 | | Stddev | 0.197199 | | Stddev | 0.018745 |
| cv | 0.012502 | | Cv | 0.273065 | | cv | 0.028991 |

Table 5. Linux comparison

| Direct | | | Boost | | | std::bitset | |
|--------|----------|---|-------|-----------|---|-------------|-----------|
| | Cpu time | | | Cpu time | | | Cpu time |
| Mean | 1.28494 | | Mean | 1.40608 | | Mean | 1.32964 |
| Median | 1.28477 | | Median | 1.40581 | | Median | 1.32958 |
| Stddev | 0.001187 | | Stddev | 0.0020054 | | Stddev | 0.0007834 |
| cv | 0.001848 | | Cv | 0.0028525 | | cv | 0.0011783 |

Table 6. Win10 comparison

| Direct | | | Boost | | | std::bitset | |
|--------|----------|---|-------|----------|---|-------------|----------|
| | Cpu time | | | Cpu time | | | Cpu time |
| Mean | 1.3221 | | Mean | 1.44832 | | Mean | 1.36632 |
| Median | 1.32597 | | Median | 1.44231 | | Median | 1.3587 |
| Stddev | 0.011807 | | Stddev | 0.014543 | | Stddev | 0.01554 |
| cv | 0.008931 | | Cv | 0.010041 | | cv | 0.011374 |

Figure 1 shows the CPU Time median and mean for all operating systems and algorithm implementations in graphical form. Notice the graphs for Windows 10 and Linux are almost identical due to the same machine being used.

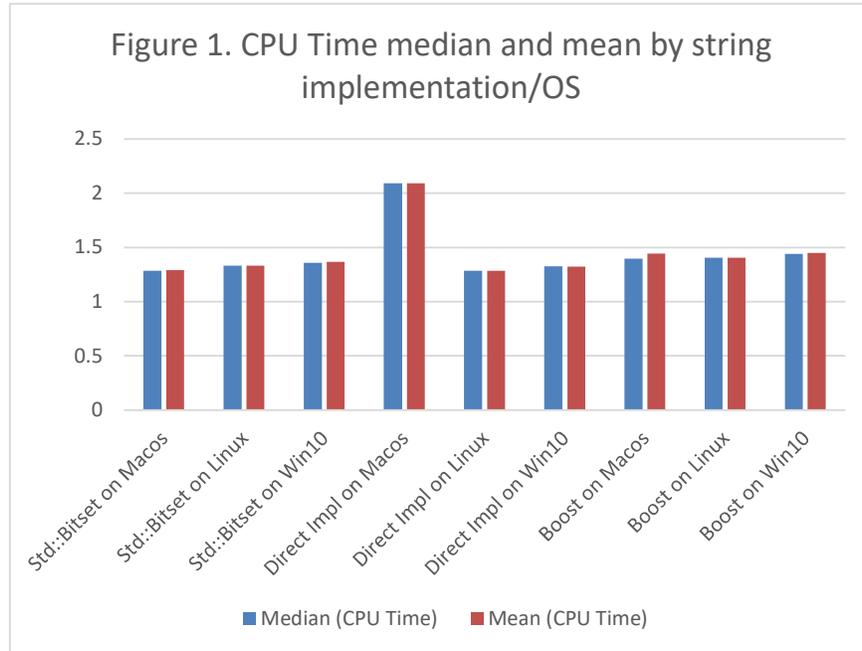

Figure 1. CPU Time median and mean by string implementation/OS

## 4    Discussion

### 4.1    Performance Analysis

For this analysis we looked at the median. On MacOS, it was clear that the fastest method was std::bitset (1.28691ms). The intermediate method was Boost (1.39833ms), whereas the slowest method was the direct implementation (2.09056ms).

On Linux, the fastest method was the direct implementation (1.28477ms). The next fastest was std::bitset (1.32958ms) and the slowest was Boost (1.40581ms). While slower, Boost was still in a decently reasonable range.

Similarly, on MYS2 (Windows 10), the fastest method was the direct implementation (1.32597ms). The second fastest method was std::bitset (1.3587ms) and the slowest was Boost (1.44231ms). The order is the same as on Linux, most likely because it is the same dual boot machine. Performance can be better due to a more efficient representation.

It is clear that the GCC optimisations on the MacBook, for that specific hardware (M2 processor) favoured the fixed-size container std::bitset implementation. On the other hand, the direct implementation underperformed significantly, which might be due to differences in memory allocation or other system-specific factors such as different CPU registers, different hardware architecture or better system implementations of std::bitset and Boost on the Mac. On the other x64 machine, the direct implementation performed better. Perhaps on x64 the two library implementations should be optimised further.

## 4.2    Cross Platform Observations

On Windows 10 (Msys2) and Linux, the direct implementation is fastest. It performed poorly, however on MacOS, possibly due to the different hardware (slower memory perhaps), or differences in memory allocation or compiler optimisations.

std::bitset performed reasonably well across all platforms, and maintained a consistent median execution time. On Macos it was the fastest and reasonably performed on Windows 10 and Linux.

While Boost was the slowest, its overtime was not excessive. It can still be a viable choice for some applications where dynamic resizing and additional features are required.

## 5    Conclusion

Given the current parameter settings, with MacOS, std::bitset offered the best performance. On Linux and Windows, the direct implementation is optimal. If flexibility is prioritised over raw performance, one can use boost::dynamic_bitset.

This study has provided useful insights for developers seeking to optimise bitstring operation performance in diverse environments.

Further research could test a std::vector<bool> implementation as mentioned by O'Dwyer[4], however there are issues with this. Usually one is recommended not to use this and stick to using std::bitset from the **STL** or boost::dynamic_bitset from the Boost Library. Indeed, in one online query on the subject, there was a recommendation to use vector<unsigned char>[2].

There may also be the possibility of creating a better algorithm instead of the direct implementation which would be more optimised and perform even better.

**Acknowledgments.** This work was supported by financial support of research project NPU I No. MSMT-7778/2014 by the Ministry of Education of the Czech Republic, by the European Regional Development Fund under the Project CEBIA-Tech No. CZ.1.05/2.1.00/03.0089 and by resources of A.I.Lab research group at Faculty of Applied Informatics, Tomas Bata University in Zlin (ailab.fai.utb.cz).

---

[2] c++ - bitwise operations on vector<bool> - Stack Overflow, at https://stackoverflow.com/questions/4008749/bitwise-operations-on-vectorbool

### REFERENCES

[1] Meli, Clyde, 'Application and Improvement of Genetic Algorithms and Genetic Programming towards the fight against Spam and other Internet Malware', PhD Thesis, University of Malta, Malta, 2017.

[2] 'Boost C++ Libraries'. Accessed: Apr. 26, 2012. [Online]. Available: http://www.boost.org/

[3] J. Merelo Guervós *et al.*, *Benchmarking Languages for Evolutionary Algorithms*. 2016, p. 41. doi: 10.1007/978-3-319-31153-1_3.

[4] A. O'Dwyer, 'Bit-vector manipulations in standard C++', Arthur O'Dwyer. Accessed: July 18, 2024. [Online]. Available: https://quuxplusone.github.io/blog/2022/11/05/bit-vectors/

[5] C. Knauth *et al.*, 'Practically efficient methods for performing bit-reversed permutation in C++11 on the x86-64 architecture', Aug. 02, 2017, *arXiv*: arXiv:1708.01873. Accessed: June 22, 2024. [Online]. Available: http://arxiv.org/abs/1708.01873

[6] 'Dr. Dobb's | CUDA, Supercomputing for the Masses: Part 1 | April 15, 2008'. Accessed: Aug. 07, 2009. [Online]. Available: http://www.ddj.com/architect/207200659

[7] P. Gustafsson, 'Programming Efficiently with Binaries and Bit Strings', in *Erlang/OTP User Conference*, 2007.

[8] V. Pieterse, D. G. Kourie, L. Cleophas, and B. W. Watson, 'Performance of C++ bit-vector implementations', in *Proceedings of the 2010 Annual Research Conference of the South African Institute of Computer Scientists and Information Technologists*, in SAICSIT '10. New York, NY, USA: Association for Computing Machinery, Oct. 2010, pp. 242–250. doi: 10.1145/1899503.1899530.

[9] Google Inc., *Google Benchmark Library*. (2021). [Online]. Available: https://github.com/google/benchmark

[10] C. Meli, V. Nezval, Z. Kominkova Oplatkova, and V. Buttigieg, 'Spam Detection Using Linear Genetic Programming', in *23rd International Conference on Soft Computing (MENDEL 2017)*, Springer, 2017.

## 6    Appendix 1

### 6.1    Hardware Used

Tech Specs:
- Desktop PC: having a Gigabyte motherboard (AMD Ryzen 7 7700X processor @4.5Ghz, 32MB of L3 cache, 1MB of L2 cache per core and 64KB of L1 cache per core) and dual-booting 64-bit Windows 10 and Void Linux. Msys2 was used on Windows 10.
- Laptop: Macbook Pro (13-inch, M2, 2022) with Apple M2 Processor @3.5Ghz running Sonoma 14.7.4 with Homebrew.